\documentclass[12pt]{article}
\usepackage[utf8]{inputenc}
\usepackage[T1]{fontenc}
\usepackage{amsmath,amssymb}
\usepackage{graphicx}
\usepackage{booktabs}
\usepackage{natbib}
\usepackage{hyperref}
\usepackage{lineno}
\usepackage{xcolor}
\usepackage{multirow}
\usepackage{caption}
\usepackage[margin=2.5cm]{geometry}
\usepackage[section]{placeins}  
\usepackage{newtxtext,newtxmath}  

\title{How many labels do you need? A decision framework for cross-habitat marine species recognition}

\author{Alzayat Saleh$^{1,2}$ \and Mostafa Rahimi Azghadi$^{1,2}$\\[6pt]
\small $^1$College of Science and Engineering, James Cook University, Townsville, QLD, Australia\\
\small $^2$Centre for AI and Data Science Innovation, James Cook University, Townsville, QLD, Australia\\
\small \texttt{alzayat.saleh@my.jcu.edu.au}}
\date{}

\begin{document}

\maketitle
\thispagestyle{empty}



\begin{abstract}
\noindent
1.~Ecological monitoring programmes increasingly rely on automated image recognition to scale beyond what human annotators can process, yet ecologists lack evidence-based guidance on how much labelling effort is required to deploy these systems reliably at new sites.

\noindent
2.~We present a decision framework, derived from a systematic cross-ecosystem evaluation, that quantifies the trade-off between labelling effort and recognition accuracy when transferring vision recognition systems across marine habitats; the benchmark spans five datasets across three oceans and three taxonomic groups (fish, corals, and invertebrates), from tropical reefs in the Great Barrier Reef and French Polynesia to a temperate Danish fjord.

\noindent
3.~We evaluated four widely available recognition models (DINOv2, CLIP, ResNet-50, and EfficientNet-B4) under four adaptation strategies (linear probing, Low-Rank Adaptation, Visual Prompt Tuning, and full fine-tuning) across three complementary protocols: within-habitat transfer across 20 reef sites (240 runs), cross-dataset geographic transfer along a difficulty gradient (40 runs), and few-shot adaptation curves with 0--100 labelled target samples per class (648 runs).

\noindent
4.~Frozen self-supervised foundation model features (DINOv2 with a linear classifier, only 1{,}538 trainable parameters) generalised to unseen reef sites at least as well as fully fine-tuned convolutional baselines four orders of magnitude larger; feature analysis showed why, with foundation models learning species-diagnostic representations that are habitat-invariant whereas convolutional baselines encode habitat-specific visual shortcuts that fail at new sites.

\noindent
5.~As few as 10--20 labelled images per species were sufficient to deploy reliable recognition at a new site, reducing annotation effort by roughly an order of magnitude compared with conventional training.

\noindent
\textbf{Solution.} Monitoring programmes expanding to new sites can deploy reliable automated species recognition by pairing a frozen, openly available foundation model (DINOv2) with a simple linear classifier and annotating only 10--20 images per species, an effort of roughly 1--4 hours per site. The decision framework presented here lets programmes budget labelling effort against expected accuracy when expanding across sites, ecosystems, or imaging platforms, and applies to any ecological monitoring programme that must generalise across sites with limited labelled data.
\end{abstract}

\vspace{0.5em}

\vspace{1.5em}

\section{Introduction}

Coral reef monitoring programmes along the Great Barrier Reef generate upwards of 100\,000 images per annual survey, each requiring expert examination to identify and count the organisms present \citep{williams2019leveraging}.
A single trained annotator processes roughly 50--100 images per hour for species-level identification, with inter-observer agreement often below 80\% even among experts \citep{thompson1997interobserver}; scaling such effort to continental monitoring programmes demands thousands of expert-hours that most research groups cannot afford \citep{beijbom2015coralnet}.
Automated species recognition based on deep learning can close this gap, and classification accuracies exceeding 90\% have been reported on individual marine datasets \citep{villon2018coral, ditria2020automating}.
Yet these systems share a critical limitation: a model trained at one site rarely works at another without extensive relabelling, and ecologists deploying automated recognition at a new site lack evidence-based guidance on a basic operational question: \emph{how many labelled images from this new site do we actually need?}

\begin{figure}[!t]
\centering
\includegraphics[width=\textwidth]{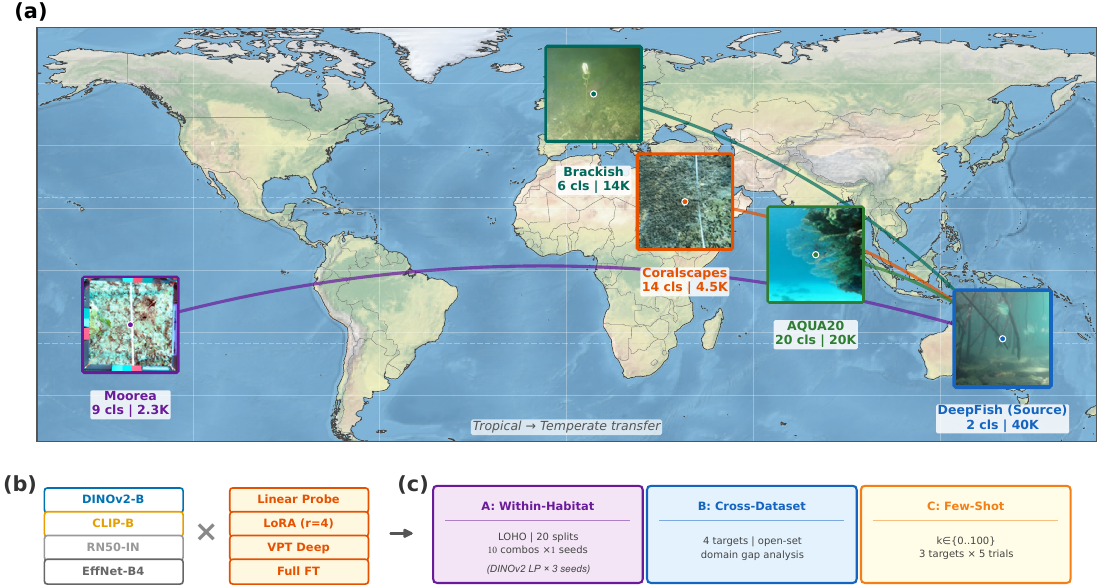}
\caption{\textbf{Experimental framework for cross-habitat marine recognition.} (a)~Geographic distribution of five marine datasets spanning tropical reefs (DeepFish, AQUA20, Moorea, Coralscapes) to temperate fjords (Brackish), shown on a satellite map with transfer arrows from the DeepFish source to four target domains. The tropical-to-temperate gradient defines an increasing domain gap. (b)~Four backbone models are crossed with four adaptation strategies to yield ten model--adaptation combinations. (c)~Three evaluation protocols test within-habitat generalisation (Protocol~A), cross-dataset geographic transfer (Protocol~B), and few-shot adaptation efficiency (Protocol~C).}
\label{fig:overview}
\end{figure}

To answer this question, we construct a benchmark that spans five marine datasets across three oceans, from tropical reefs to temperate fjords, and systematically evaluate how modern vision foundation models transfer across these habitats (Figure~\ref{fig:overview}).


\subsection*{The Cross-Habitat Transfer Challenge in Marine Systems}

Underwater imagery is collected at scale through diverse platforms: baited remote underwater video stations (BRUVS), diver-operated cameras, towed camera sleds, remotely operated vehicles, and fixed camera arrays \citep{mallet2014underwater}.
Coral reef benthic surveys typically photograph quadrats at fixed transect points, generating hundreds of images per dive site \citep{beijbom2012moorea}.
The bottleneck is not collection but annotation, and existing automated tools such as CoralNet require site-specific training sets of several thousand labelled points before achieving acceptable accuracy \citep{beijbom2015coralnet}.

Cross-habitat transfer compounds the problem.
Monitoring programmes span sites that differ in depth, turbidity, substrate, and species composition, and a classifier trained on clear tropical reef imagery performs poorly when deployed in a turbid temperate estuary \citep{durden2016perspectives}.
The model learns habitat-specific shortcuts (water colour, background texture) rather than species-diagnostic features (body shape, fin configuration, skeletal structure), and these shortcuts fail at new sites.
Four properties of underwater imagery make this transfer particularly challenging for computer vision: (i) wavelength-dependent light attenuation distorts colour as a function of depth and water type; (ii) suspended particles reduce contrast and obscure fine detail; (iii) many marine organisms are camouflaged against their substrate; and (iv) species abundance distributions are highly skewed, with a few common species and many rare ones \citep{durden2016perspectives}.
These properties motivate models that rely on structural and textural features rather than colour, and that can adapt to new visual domains from small numbers of examples.

\subsection*{Foundation Models and Parameter-Efficient Adaptation}

Vision foundation models offer a potential answer.
These are large neural networks pretrained on hundreds of millions of images, either from unlabelled images alone (self-supervised learning) or from images paired with text descriptions (language-supervised learning), producing general-purpose visual representations that can be adapted to new tasks with minimal additional data \citep{oquab2024dinov2, radford2021clip}.
The analogy to ecological expertise is instructive: an experienced marine biologist arriving at an unfamiliar reef can identify species far more quickly than a novice, because years of field experience have built an internal representation of what distinguishes one species from another, independent of site-specific conditions.
Foundation models acquire a similar ``visual experience'' during pretraining, learning to encode shape, texture, and structure in ways that generalise across visual domains.
Domain-specific adaptations have followed for ecology: BioCLIP \citep{stevens2024bioclip} fine-tunes a CLIP backbone (the underlying pretrained network) on 10 million biodiversity images spanning 450{,}000+ species, and FathomNet \citep{katija2022fathomnet} provides over 100{,}000 annotated underwater images that could serve as a marine pretraining resource, but no marine-specific foundation model has yet been published.

Adapting a pretrained foundation model to a new monitoring site can take several forms.
The simplest is linear probing, in which the backbone is frozen and only a single classifier layer is trained on top of the extracted features.
Parameter-efficient fine-tuning (PEFT) methods modify a small fraction of the model's parameters: Low-Rank Adaptation \citep{hu2022lora} inserts low-rank matrices into the attention layers, and Visual Prompt Tuning \citep{jia2022vpt} prepends learnable tokens at every transformer layer.
Full fine-tuning, in which all parameters are updated, remains the conventional approach in marine computer vision \citep{saleh2022fish_survey, gonzalez2020monitoring} but typically requires large labelled datasets and is prone to overfitting when target data are scarce.
Despite the ecological appeal of these methods, prior work in marine computer vision has largely treated each dataset as an independent problem \citep{saleh2020deepfish, gonzalez2020monitoring}, and cross-site transfer has received limited systematic attention; studies that do evaluate generalisation typically consider a single model on a single dataset pair \citep{christin2019applications, weinstein2018computer}.
Three gaps remain: (a) no multi-dataset benchmark exists for evaluating cross-habitat transfer in marine systems; (b) modern foundation models and parameter-efficient adaptation methods have not been systematically compared for underwater imagery; and (c) ecologists lack evidence-based guidance on the number of labelled samples required at a new site.

This paper addresses all three gaps through the following contributions:

\begin{enumerate}
    \item A \emph{decision framework for ecological monitoring} that quantifies the trade-off between labelling effort and recognition accuracy at new sites, derived from systematic evaluation across five marine ecosystems spanning three oceans and three taxonomic groups (fish, corals, and invertebrates).
    \item A \emph{cross-ecosystem evaluation methodology} based on an explicit difficulty gradient (within-habitat $\rightarrow$ cross-dataset $\rightarrow$ cross-taxon), allowing monitoring programmes to anticipate how recognition performance degrades when transferring across geographic and ecological boundaries.
    \item \emph{Evidence-based labelling guidelines} showing that 10--20 labelled images per species suffice for reliable deployment at a new site, reducing annotation effort by roughly an order of magnitude compared to conventional training-from-scratch.
\end{enumerate}

Although we demonstrate the framework on marine datasets, the evaluation protocols and adaptation methodology apply to any ecological monitoring system that must generalise across sites, seasons, or imaging platforms; camera trap networks, acoustic monitoring arrays, and aerial survey programmes face the same cross-site transfer challenge \citep{tuia2022perspectives, schneider2019past}.


\section{Materials and Methods}

\subsection{Study System: A Five-Dataset Marine Benchmark}

\begin{figure}[!htbp]
\centering
\includegraphics[width=0.92\textwidth]{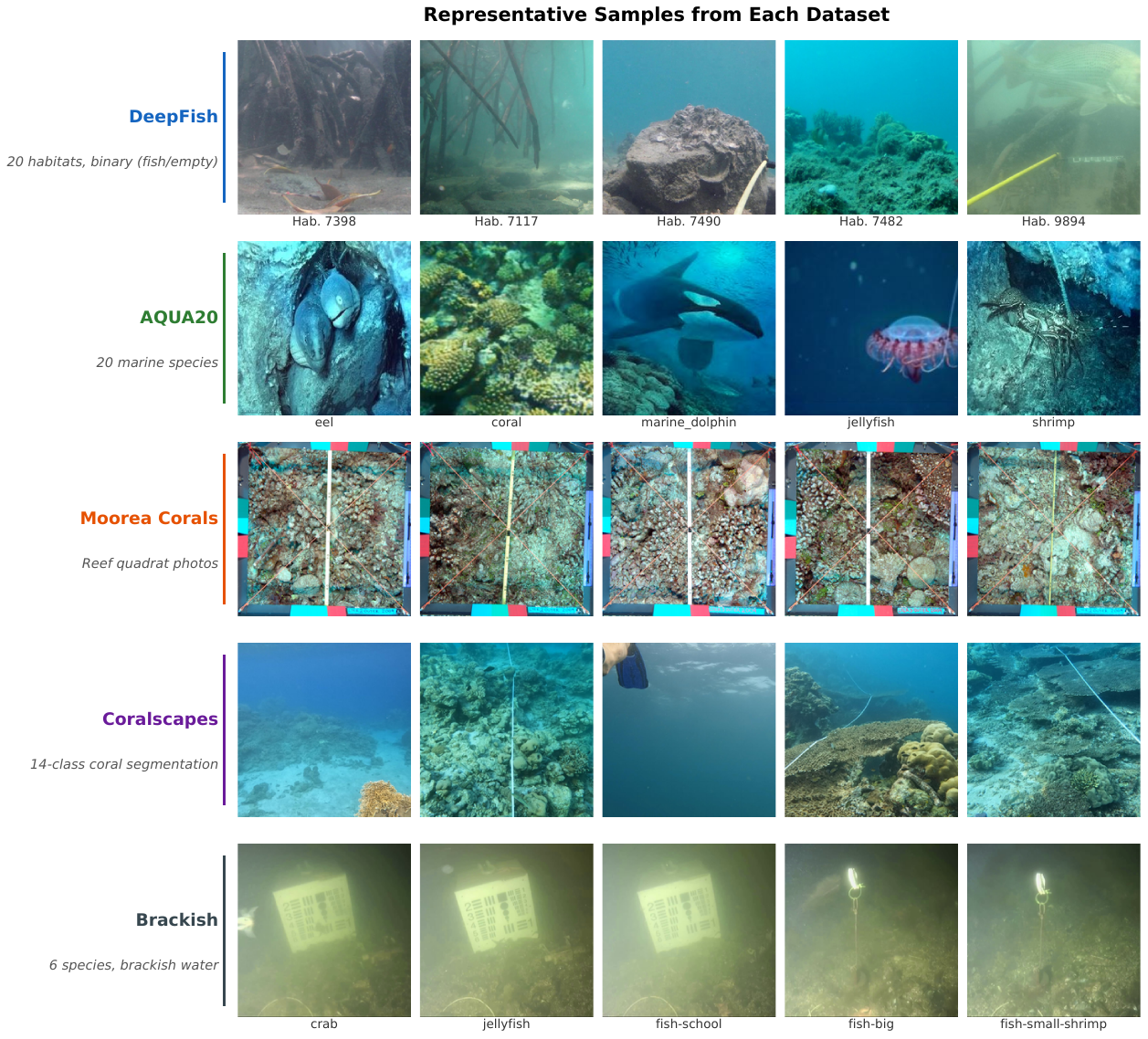}
\caption{Representative images from each dataset. Each row shows five randomly sampled images from different habitats or classes within a single dataset. The visual domain gap between datasets (differences in water colour, turbidity, lighting, and substrate) is the core challenge this work addresses.}
\label{fig:samples}
\end{figure}

We assembled a benchmark of five publicly available marine image datasets, selected to span a gradient of ecological and visual diversity (Table~\ref{tab:datasets}; Figure~\ref{fig:samples}). All datasets were collected and released by their respective original authors under the permits, ethical approvals, and open licences described in the corresponding publications cited below; the present study performed no primary field data collection.

\textbf{DeepFish} \citep{saleh2020deepfish} comprises 39\,766 images from 20 distinct reef habitats along the Great Barrier Reef, Australia.
Each image is labelled as containing fish (valid) or empty.
The 20 habitats vary in depth, substrate, and turbidity, making this dataset uniquely suited to within-habitat transfer experiments where species labels are shared but visual conditions differ.

\textbf{AQUA20} \citep{fuad2025aqua20} contains 8\,171 images of 20 marine species, drawn from mixed tropical and subtropical sources.
Species range from reef fish to cephalopods, providing a multi-class classification benchmark with moderate taxonomic breadth.
Class sizes in the training split range from 20 to 1\,329 images, with six classes containing fewer than 50 samples.

\textbf{Moorea Corals} \citep{beijbom2012moorea, edmunds2019mcrlter} includes approximately 2\,050 benthic quadrat images from the Moorea Coral Reef Long Term Ecological Research site in French Polynesia, with point annotations for nine substrate classes: crustose coralline algae (CCA), \emph{Montipora} (MONT), \emph{Porites} (POR), \emph{Pocillopora} (POCILL), \emph{Montastraea} (MCAV), turf algae (TURF), macroalgae (MACRO), sand (SAND), and ``other'' (OTHER).
We converted point annotations to image-level classification labels using the majority class per image.
The resulting class distribution is highly skewed: the dominant class (CCA) has 515 training images while three classes have fewer than 20 (POCILL: 14, TURF: 17, MCAV: 18), and one class (POCILL) has zero test samples.

\textbf{Coralscapes} \citep{sauder2025coralscapes} provides approximately 2\,075 high-resolution images from Red Sea reefs with pixel-level semantic segmentation masks for 39 benthic classes.
We derived image-level labels by assigning each image to its dominant class, yielding 14 classes with sufficient representation.
The resulting distribution is severely long-tailed: the largest class has 458 training images while eight classes have fewer than 10 samples, making Coralscapes the most imbalanced target in our benchmark.

\textbf{Brackish} \citep{pedersen2019brackish} contains approximately 14\,500 frames extracted from 89 underwater videos recorded in Limfjorden, a cold temperate fjord in Denmark.
Six species are annotated with bounding boxes; we converted detections to image-level presence labels.
All six classes are well-represented, with training samples ranging from 161 to 5\,948.
This dataset represents the most visually distinct domain in our benchmark: murky, greenish water with low contrast, entirely different from tropical reef imagery.

The five datasets define a transfer difficulty gradient anchored by ecological and geographic distance:

\begin{itemize}
    \item \textbf{Easiest:} Within DeepFish (habitat A to habitat B within the same reef system)
    \item \textbf{Easy:} DeepFish to AQUA20 (tropical fish to mixed marine species, same taxon)
    \item \textbf{Medium:} DeepFish to Brackish (tropical reef to cold temperate fjord, same taxon but different biome)
    \item \textbf{Hard:} DeepFish to Moorea (fish to coral, cross-taxon and cross-ocean)
    \item \textbf{Hardest:} DeepFish to Coralscapes (fish to coral, cross-taxon with severe class imbalance)
\end{itemize}

This gradient is a core organising principle of our experimental design: it allows us to measure how gracefully different models degrade as the visual and ecological gap between source and target widens.

\begin{table}[t]
\centering
\caption{Summary of the five marine datasets. The transfer difficulty gradient increases from top to bottom.}
\label{tab:datasets}
\small
\resizebox{\textwidth}{!}{%
\begin{tabular}{@{}llrrrlll@{}}
\toprule
Dataset & Region & Images & Classes & Min/Max cls. & Task & Platform & Ecological Application \\
\midrule
DeepFish       & GBR, Australia     & 39\,766 & 2  & 7\,654/12\,112 & Classification & Baited video  & Multi-habitat reef fish monitoring \\
AQUA20         & Mixed tropical     & 8\,171  & 20 & 20/1\,329   & Classification & Mixed         & Species-level identification \\
Brackish       & Limfjorden, Denmark & 14\,500 & 6  & 161/5\,948  & Classification & Fixed camera  & Cold temperate monitoring \\
Moorea Corals  & French Polynesia   & 2\,050  & 9  & 14/515      & Classification & Diver camera  & Coral reef health assessment \\
Coralscapes    & Red Sea            & 2\,075  & 14 & 1/458       & Classification & ROV/diver     & Benthic substrate mapping \\
\bottomrule
\end{tabular}%
}
\end{table}

\subsection{Vision Foundation Models and CNN Baselines}

We evaluate two categories of model: foundation models pretrained on large-scale data with self-supervised or language-supervised objectives, and convolutional neural network (CNN) baselines pretrained with supervised classification on ImageNet \citep{deng2009imagenet}.

\textbf{Foundation models.}
DINOv2-Base \citep{oquab2024dinov2} is a Vision Transformer (ViT-B/14) trained by self-distillation on 142 million curated images.
Self-supervised pretraining works by presenting the model with different augmented views of the same image and training it to produce consistent representations; the model learns to encode visual structure without any labels.
The resulting features capture shape, texture, and spatial arrangement in ways that transfer across visual domains, much as an experienced field biologist recognises species by diagnostic morphological features regardless of the site visited.
CLIP-Base \citep{radford2021clip} is a ViT-B/16 trained to match images with their text descriptions across 400 million image-text pairs, learning visual concepts grounded in natural language.
CLIP's language supervision provides a complementary learning signal: the model learns that images described with the same words should have similar representations, enabling zero-shot classification, in which images are assigned to categories from a text description of each category alone, with no labelled training examples.

\textbf{CNN baselines.}
ResNet-50 \citep{he2016resnet} and EfficientNet-B4 \citep{tan2019efficientnet} are convolutional architectures pretrained on ImageNet-1K (1.28 million images, 1\,000 classes) with supervised classification.
These baselines represent the standard approach in applied marine computer vision, where models are initialised from ImageNet weights and fine-tuned on task-specific data.
We include both architectures to control for the possibility that performance differences stem from the transformer architecture itself rather than the pretraining strategy.

Table~\ref{tab:models} summarises the model specifications.
We note that EfficientNet-B4 uses a native input resolution of $380 \times 380$ (2.88$\times$ more pixels than the $224 \times 224$ used by the other three models), which may confer an advantage from higher-resolution input rather than architectural or pretraining differences alone.

\begin{table}[t]
\centering
\caption{Model specifications. Foundation models use self-supervised or language-supervised pretraining; CNN baselines use supervised ImageNet pretraining.}
\label{tab:models}
\small
\begin{tabular}{@{}lllcrr@{}}
\toprule
Model & Architecture & Pretraining & Image Size & Feature Dim & Parameters \\
\midrule
DINOv2-Base & ViT-B/14 & Self-supervised (142M images)        & $224 \times 224$ & 768  & $\sim$86M \\
CLIP-Base   & ViT-B/16 & Language-image (400M pairs)           & $224 \times 224$ & 512  & $\sim$86M \\
ResNet-50   & ResNet-50 & Supervised ImageNet-1K               & $224 \times 224$ & 2048 & $\sim$24M \\
EfficientNet-B4 & EfficientNet-B4 & Supervised ImageNet-1K     & $380 \times 380$ & 1792 & $\sim$18M \\
\bottomrule
\end{tabular}
\end{table}

\subsection{Adaptation Strategies}

Rather than updating all parameters in a pretrained model (which risks overfitting when labelled data are scarce), parameter-efficient adaptation methods modify only a small fraction of the model while keeping the pretrained backbone frozen.
We employ model-dependent adaptation strategies: transformer-specific methods (LoRA and VPT) for the Vision Transformer foundation models, and full fine-tuning for the CNN baselines.
All models are additionally evaluated with a linear probe as the simplest baseline.

\textbf{Linear Probe} (all models).
The backbone is entirely frozen and a single linear classification layer is trained on top of the extracted features.
This tests the quality of the pretrained representations without any adaptation; analogously, it asks the model questions about what it already knows, without teaching it anything new.
The classification head has on the order of thousands of parameters.

\textbf{Low-Rank Adaptation} (LoRA; foundation models only).
LoRA \citep{hu2022lora} inserts small, low-rank matrices into the attention layers of a Vision Transformer, allowing targeted adjustments to how the model attends to visual features.
We apply LoRA with rank $r = 4$, scaling factor $\alpha = 8$, and dropout 0.1 to the query and value projections.
This adds approximately 149K trainable parameters for DINOv2-Base (0.17\% of the full model) and 75K for CLIP-Base (0.09\%), yet enables meaningful adaptation of the attention mechanism.

\textbf{Visual Prompt Tuning} (VPT-Deep; foundation models only).
VPT \citep{jia2022vpt} prepends a set of learnable ``prompt'' tokens at every transformer layer, providing the model with task-specific context at each processing stage.
We use 10 prompt tokens per layer with dropout 0.1, adding approximately 94K trainable parameters for DINOv2-Base (0.11\%) and 93K for CLIP-Base (0.06\%).

\textbf{Full Fine-tuning} (CNN baselines only).
All parameters are unfrozen and trained end-to-end.
This is the conventional approach in applied marine computer vision, and it represents the upper bound of what can be achieved by updating the entire model.
We apply layer-wise learning rate decay (factor 0.75) and label smoothing (0.1) to mitigate overfitting.

All configurations use the AdamW optimiser \citep{loshchilov2019adamw} with cosine learning rate scheduling, 5\% linear warmup, early stopping with patience of 10 epochs, and mixed-precision training.
Table~\ref{tab:adaptations} summarises the adaptation strategies and their key hyperparameters.

\begin{table}[t]
\centering
\caption{Adaptation strategy summary. Trainable parameter counts are approximate and vary by model. Foundation models (DINOv2-B, CLIP-B) use Linear Probe, LoRA, and VPT; CNN baselines (ResNet-50, EfficientNet-B4) use Linear Probe and Full Fine-tuning. CLIP's text encoder (63.4M parameters) is loaded to support zero-shot classification (Protocol~C, $k=0$) but is excluded from the training optimiser in all adaptation configurations; the trainable parameter counts reported here reflect only the visual encoder adaptation and classification head. Training times for CLIP linear probe (${\sim}5$ minutes per run) are consistent with other linear probe configurations, confirming that the text encoder is not updated during training.}
\label{tab:adaptations}
\small
\begin{tabular}{@{}lllrl@{}}
\toprule
Strategy & Applies To & Trainable Params (\% of total) & LR & Epochs \\
\midrule
Linear Probe    & All models       & $\sim$0.01\% & $1 \times 10^{-3}$ & 50 \\
LoRA ($r=4$)    & Foundation models & $\sim$0.1--0.2\% & $1 \times 10^{-4}$ & 30 \\
VPT-Deep        & Foundation models & $\sim$0.1\%  & $1 \times 10^{-4}$ & 30 \\
Full Fine-tune  & CNN baselines    & 100\%          & $5 \times 10^{-5}$ & 20 \\
\bottomrule
\end{tabular}
\end{table}

\subsection{Experimental Protocols}

We designed three complementary protocols, each answering a distinct question about transferability.
Together, they trace a gradient from the easiest transfer scenario (same reef system, different habitat) to the hardest (cross-taxon transfer to severely imbalanced coral datasets with few labelled samples).

\subsubsection{Protocol A: Can Foundation Models Recognise Species in a Reef Habitat They Have Never Seen?}

Binary species occupancy detection, distinguishing images that contain target organisms from empty frames, is a foundational task in underwater monitoring.
BRUVS programmes routinely process thousands of video frames to identify which contain fish \citep{saleh2020deepfish}, and automated occupancy detection is the first stage in multi-step monitoring pipelines.
Using the 20 habitats in DeepFish, we perform leave-one-habitat-out evaluation: train on 19 habitats, evaluate on the held-out habitat, and repeat for all 20.
The controlled nature of this binary protocol (shared label space, identical annotation procedure, 20 visually distinct habitats) makes it uniquely suited to isolating the effect of habitat-induced domain shift without confounding factors such as differing label spaces or annotation inconsistencies.
Each of the 10 model--adaptation configurations is evaluated across all 20 target habitats with a single seed (seed = 42), yielding 200 runs.
To assess sensitivity to random initialisation, we additionally ran DINOv2 LP (the top-performing configuration) with two further seeds (123, 456) across all 20 habitats, yielding 240 total Protocol~A runs.
We report macro F1, a recognition-accuracy score from 0 (worst) to 1 (best) that weights all classes equally, averaged across the 20 held-out habitats, with standard deviation capturing inter-habitat variability.

\subsubsection{Protocol B: How Far Can Representations Transfer Across Marine Ecosystems?}

We evaluate cross-dataset transfer along the difficulty gradient defined in Section~2.1, using DeepFish as the source for all four transfer pairs.
Because source and target datasets have non-overlapping label spaces, we use open-set evaluation based on nearest-prototype matching in feature space: each target class is represented by the mean feature vector of its labelled examples, and test images are assigned to the nearest prototype.
This approach requires no shared labels and is therefore always applicable.
We note that nearest-prototype matching inherently favours models with better-organised feature spaces; we select this method because it is the simplest non-parametric classifier applicable to open-set transfer and because it mirrors the practical scenario where an ecologist has labelled examples of target species but no target-domain training infrastructure.
Each of the 10 model--adaptation configurations is evaluated on all four pairs, yielding 40 runs.

\subsubsection{Protocol C: How Many Labelled Samples from a New Site Are Actually Needed?}

This is the central experiment.
For three transfer pairs spanning the difficulty gradient (DeepFish to AQUA20, DeepFish to Moorea, DeepFish to Brackish), we vary the number of labelled target samples per class ($k$) from 0 to 100 ($k \in \{0, 1, 2, 5, 10, 20, 50, 100\}$) and measure performance as a function of $k$.
At each $k$ value, we draw five independent random samples of target data for $k \leq 20$ and three for $k \geq 50$, reporting mean $\pm$ standard deviation across trials (at $k = 0$ no sampling is required, so the result is deterministic).
Training uses a combined strategy: source data and $k$ target samples per class are mixed and the model is trained on the combined set.
For CLIP at $k = 0$, we additionally evaluate zero-shot classification using text prompts of the form ``a photo of a \{species\} in the ocean''.
Six model--adaptation configurations are evaluated: DINOv2-B with linear probe, DINOv2-B with LoRA ($r=4$), CLIP-B with linear probe, CLIP-B with LoRA ($r=4$), ResNet-50 with linear probe, and EfficientNet-B4 with linear probe.
This yields 648 runs across all $k$ values, pairs, and trials.

An important practical constraint is that target datasets have uneven class distributions.
When $k$ exceeds the number of available training samples for a class, all samples from that class are assigned to the support set, leaving zero query samples for that class during evaluation.
AQUA20 is well-balanced (minimum 20 samples per class), so all $k$ values up to 20 are fully supported.
Moorea has three classes with fewer than 20 training samples (POCILL: 14, TURF: 17, MCAV: 18), so at $k = 20$ these classes are exhausted.
Brackish is the most balanced target (minimum 161 samples per class), with no sampling constraints at any $k$ value.
We report results only over classes that retain at least one query sample at each $k$, and note the effective number of evaluable classes where relevant.

\subsubsection{Metrics and Statistical Analysis}

We use macro F1 as the primary metric for Protocols A and B because it handles class imbalance by weighting each class equally, which is essential for datasets with skewed species abundances.
For Protocol~C, we report balanced accuracy because the few-shot sampling protocol progressively depletes the query set at high $k$, causing macro F1 to decline artifactually as minority classes are exhausted from evaluation.
For cross-protocol comparability, we report both metrics in Appendix~S1: Table~S2.
Pairwise model comparisons across habitats (Protocol A) use the Wilcoxon signed-rank test (Table~\ref{tab:wilcoxon}).
For all protocols, 10\% of the source training data is randomly held out as a validation set for early stopping.
All random seeds are fixed and all configurations specified in YAML files to ensure full reproducibility.
Code is available at \url{https://github.com/alzayats/cross-habitat-marine}.


\section{Results}

\subsection{Within-Habitat Transfer: Do Foundation Models Generalise Across Reef Sites?}

\begin{figure}[!htbp]
\centering
\includegraphics[width=0.92\textwidth]{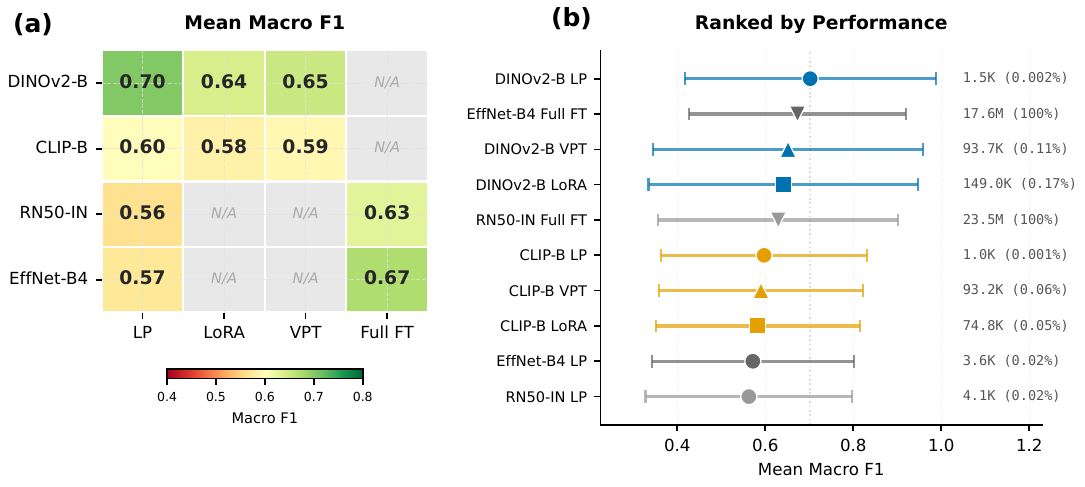}
\caption{Within-habitat transfer performance (Protocol A). (a) Heatmap showing macro F1 for each model--adaptation combination, averaged across 20 leave-one-habitat-out experiments. Grey cells indicate inapplicable combinations (e.g., LoRA is transformer-specific and does not apply to CNNs). (b) Forest plot ranking all combinations by mean F1 $\pm$ SD, with trainable parameter counts annotated. DINOv2-Base with a frozen linear probe achieves the highest mean F1 (0.70) using only 1{,}538 trainable parameters.}
\label{fig:heatmap_within}
\end{figure}

Protocol A tests whether models learn species-diagnostic features that transfer across the 20 reef habitats in DeepFish, or whether they rely on habitat-specific visual shortcuts.
Figure~\ref{fig:heatmap_within} presents the macro F1 scores for each model--adaptation combination, averaged across all 20 leave-one-habitat-out experiments.

DINOv2-Base with a linear probe achieved the highest mean macro F1 ($0.70 \pm 0.28$ across all 20 habitats; $0.74 \pm 0.23$ excluding habitat 7268, a degenerate case with zero empty-class test samples), followed by EfficientNet-B4 with full fine-tuning ($0.67 \pm 0.24$), DINOv2-Base with VPT ($0.65 \pm 0.30$), and DINOv2-Base with LoRA ($0.64 \pm 0.30$).
CLIP-Base configurations performed below, with its best adaptation (linear probe) reaching a mean F1 of $0.60 \pm 0.23$.
Among CNN baselines, EfficientNet-B4 with full fine-tuning achieved $0.67 \pm 0.24$ and ResNet-50 with full fine-tuning reached $0.63 \pm 0.27$.

Wilcoxon signed-rank tests across the 20 habitats (Table~\ref{tab:wilcoxon}) revealed that DINOv2 LP was significantly better than all three CLIP configurations ($P < 0.05$), both CNN linear probes ($P < 0.05$), and ResNet-50 with full fine-tuning ($P = 0.033$).
However, DINOv2 LP was not significantly different from DINOv2 LoRA ($P = 0.11$), DINOv2 VPT ($P = 0.29$), or EfficientNet-B4 with full fine-tuning ($P = 0.41$).
The practical implication is that no statistically significant difference was detected between DINOv2 LP and the best alternatives, yet DINOv2 LP uses only 1{,}538 trainable parameters, four orders of magnitude fewer than LoRA (149K) or full fine-tuning (17.6M); exact parameter counts for all ten configurations are reported in Appendix~S1: Table~S1.

That a frozen linear probe matches LoRA and VPT is confirmed by the statistical tests: the pretrained DINOv2 features already contain sufficient species-relevant information that parameter-efficient adaptation of the attention layers provides no measurable benefit for within-habitat transfer.
Training dynamics confirm this: DINOv2 linear probe converges in 2--3 epochs, while adapted configurations require substantially more training without improving final performance (Appendix~S1: Figure~S1).
This has practical implications: ecologists need only train a lightweight classifier head, avoiding the complexity and hyperparameter tuning of LoRA or VPT.

Across three random seeds, DINOv2 LP mean F1 varied by only 0.015 (0.70, 0.72, 0.71 for seeds 42, 123, 456 respectively), confirming that inter-habitat variability (SD $= 0.28$) dominates over initialisation variance.

\begin{figure}[!htbp]
\centering
\includegraphics[width=0.92\textwidth]{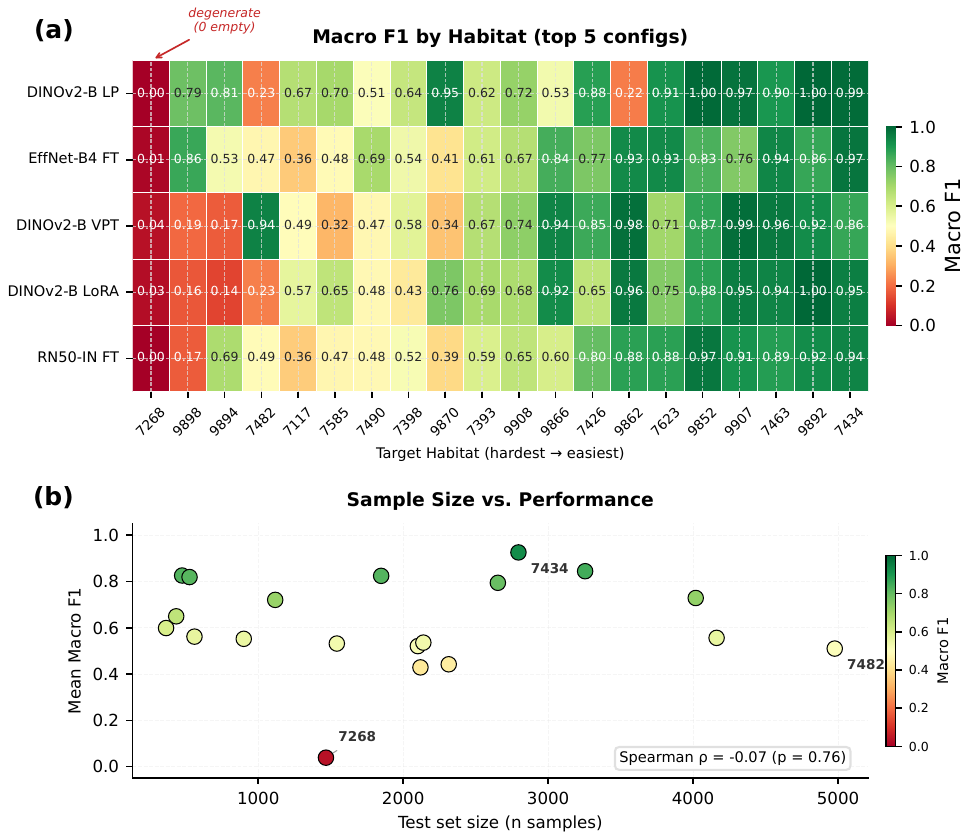}
\caption{Habitat difficulty landscape (Protocol A). (a) Heatmap of macro F1 for the top five model--adaptation combinations across all 20 habitats, sorted by difficulty (hardest left; habitats ranked by ascending mean macro F1 across the top five configurations). Habitat 7268 is degenerate (zero empty-class test samples). (b) Scatter plot of test set size vs.\ mean F1 per habitat, coloured by macro F1 (same scale as panel a). Difficulty is driven primarily by class imbalance rather than sample size (Spearman $\rho$ annotated).}
\label{fig:habitat_landscape}
\end{figure}

\begin{figure}[!htbp]
\centering
\includegraphics[width=0.92\textwidth]{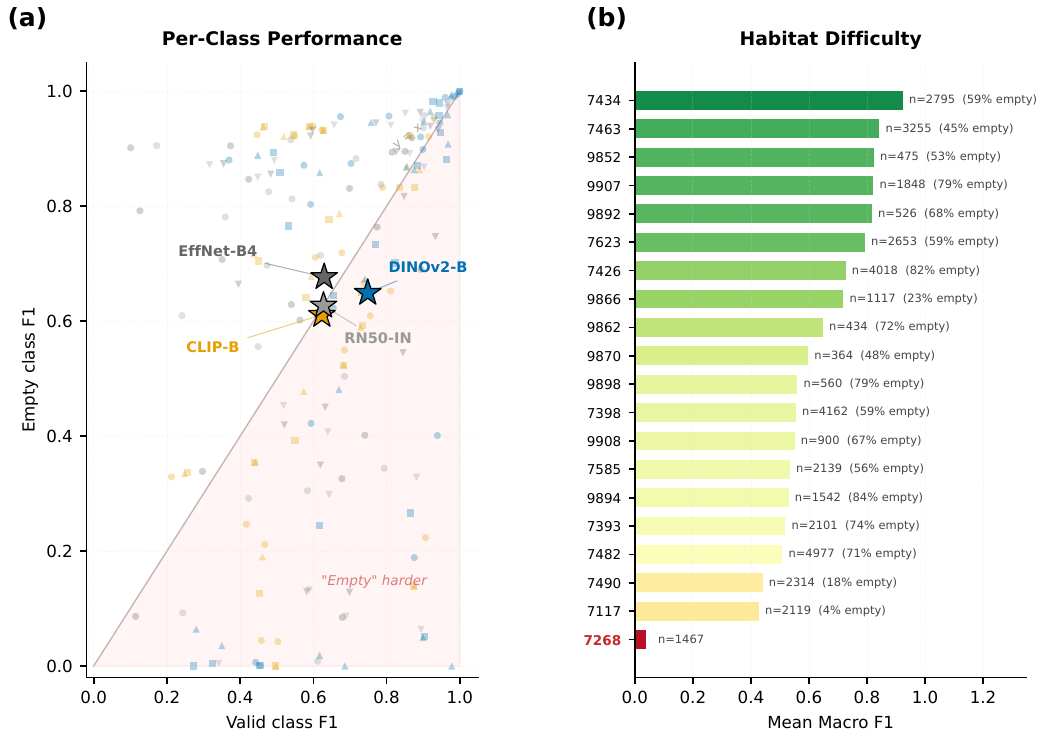}
\caption{Class imbalance analysis (Protocol A). (a) Empty-class F1 vs.\ valid-class F1 for all runs (excluding degenerate habitat 7268). Points coloured by model, shaped by adaptation. All points fall below the diagonal, indicating the ``empty'' class is systematically harder. (b) Per-habitat difficulty bar chart, sorted worst to best, with class imbalance and sample count annotated.}
\label{fig:class_imbalance}
\end{figure}

Performance varied substantially across target habitats (Figure~\ref{fig:habitat_landscape}).
The hardest habitats to generalise to were 7268 (F1 = 0.00, a degenerate case with zero empty-class test samples), 7482 (F1 = 0.23), and 9862 (F1 = 0.25), which share high class imbalance.
The easiest habitats (9892, F1 = 1.00; 9852, F1 = 0.99; 7434, F1 = 0.99) had more balanced class distributions.
This variability (standard deviation of 0.28 across habitats for DINOv2 linear probe) underscores that even within a single reef system, some sites pose greater transfer challenges than others, driven primarily by class imbalance rather than visual difficulty (Figure~\ref{fig:class_imbalance}).

For monitoring programmes operating across multiple sites within a reef system, these results indicate that a DINOv2-based model trained on a subset of sites will perform reliably at held-out sites without site-specific retraining, using only a frozen feature extractor with a linear classifier.

\begin{table}[htbp]
\centering
\caption{Wilcoxon signed-rank tests comparing DINOv2 LP (the top-performing configuration) against all other configurations across 20 habitats (Protocol~A). Significance: $^{**}P < 0.01$, $^{*}P < 0.05$, n.s.\ = not significant.}
\label{tab:wilcoxon}
\small
\begin{tabular}{@{}lccrl@{}}
\toprule
Configuration & Mean F1 & Mean Diff & $P$-value & Sig. \\
\midrule
DINOv2-B LP (ref.)     & $0.70 \pm 0.28$ & ---    & ---    & --- \\
DINOv2-B LoRA          & $0.64 \pm 0.30$ & $+0.06$ & 0.112  & n.s. \\
DINOv2-B VPT           & $0.65 \pm 0.30$ & $+0.05$ & 0.294  & n.s. \\
EffNet-B4 FT           & $0.67 \pm 0.24$ & $+0.03$ & 0.409  & n.s. \\
RN50 FT                & $0.63 \pm 0.27$ & $+0.07$ & 0.033  & * \\
CLIP-B LP              & $0.60 \pm 0.23$ & $+0.10$ & 0.011  & * \\
CLIP-B VPT             & $0.59 \pm 0.23$ & $+0.11$ & 0.007  & ** \\
CLIP-B LoRA            & $0.58 \pm 0.23$ & $+0.12$ & 0.006  & ** \\
EffNet-B4 LP           & $0.57 \pm 0.22$ & $+0.13$ & 0.044  & * \\
RN50 LP                & $0.56 \pm 0.23$ & $+0.14$ & 0.014  & * \\
\bottomrule
\end{tabular}
\end{table}

\subsection{Cross-Dataset Transfer: From Tropical Reefs to Temperate Fjords}

\begin{figure}[!htbp]
\centering
\includegraphics[width=0.92\textwidth]{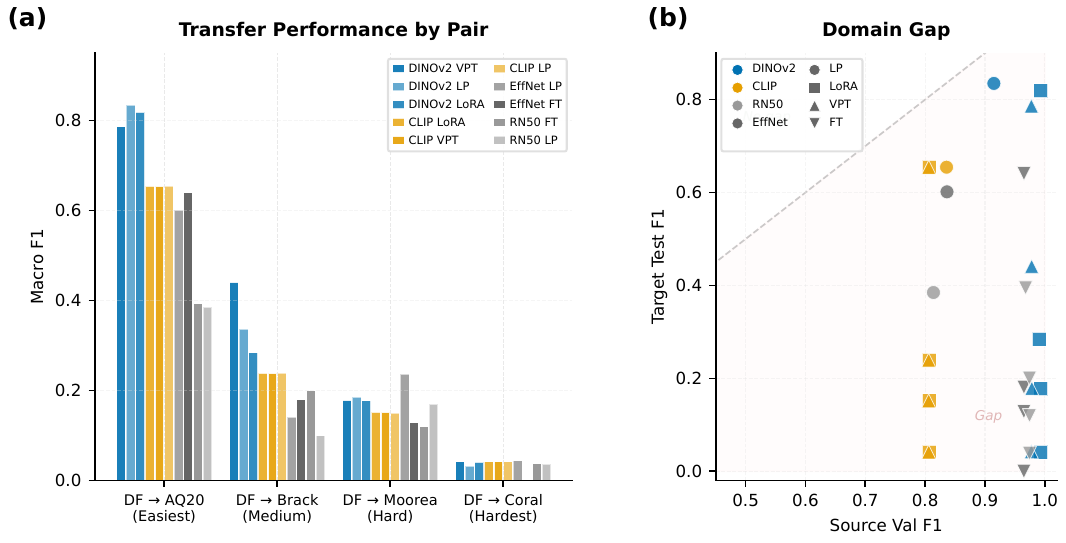}
\caption{Cross-dataset transfer performance (Protocol B). (a) Macro F1 for all model--adaptation combinations across four cross-dataset transfer pairs ordered by difficulty. (b) Domain gap scatter: source validation F1 vs.\ target test F1 per run; colour indicates model backbone and marker shape indicates adaptation strategy. All points fall well below the diagonal, indicating substantial domain shift.}
\label{fig:crossdataset_bars}
\end{figure}

Protocol B tests how far pretrained representations can transfer across datasets that differ not only in visual domain but in species composition, geographic origin, and ecological context.
Figure~\ref{fig:crossdataset_bars} presents the macro F1 for all model--adaptation combinations across the four cross-dataset transfer pairs.
Performance declined sharply along the difficulty gradient.
The easiest transfer pair (DeepFish to AQUA20) yielded the highest F1 for all models, with DINOv2 linear probe reaching 0.84 and DINOv2 LoRA reaching 0.82.
The medium-difficulty transfer (DeepFish to Brackish, same taxon but different biome) saw DINOv2 VPT reach 0.44 and DINOv2 LP reach 0.34.
The cross-taxon transfers proved substantially harder: DeepFish to Moorea yielded F1 = 0.18 for the best models, while DeepFish to Coralscapes reduced all models to near-chance performance ($\sim$0.04 F1), indicating that open-set nearest-prototype matching cannot bridge the gap between fish and coral class distributions combined with severe class imbalance.

\begin{figure}[!htbp]
\centering
\includegraphics[width=0.92\textwidth]{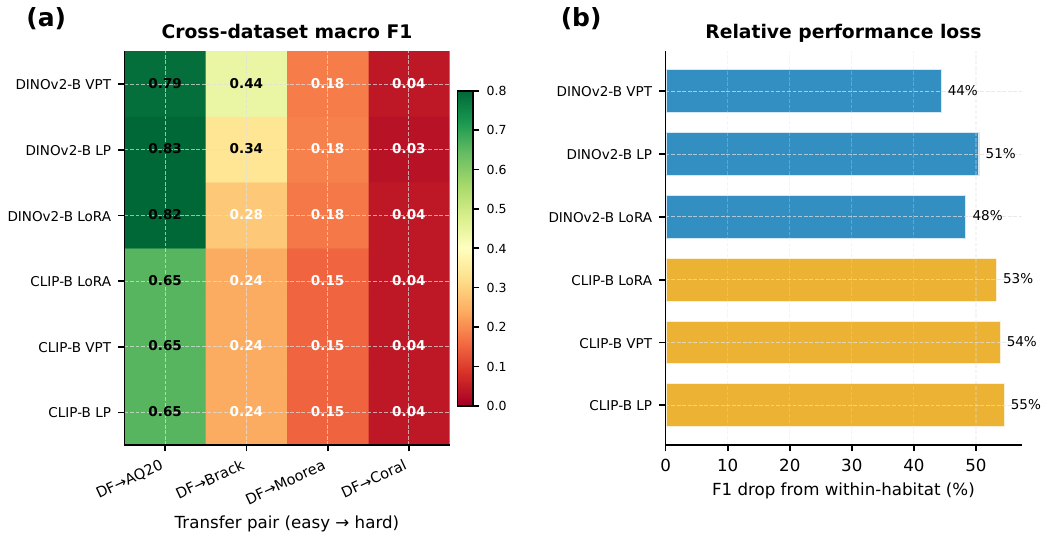}
\caption{Transfer robustness (Protocol B). (a) Heatmap of macro F1 for the top six model--adaptation combinations across all four transfer pairs, showing the monotonic decline along the difficulty gradient. (b) Relative F1 drop from within-habitat (Protocol A) baselines, quantifying how much each combination loses under domain shift.}
\label{fig:transfer_robustness}
\end{figure}

Foundation models degraded more gracefully than CNN baselines across the difficulty gradient (Figure~\ref{fig:transfer_robustness}).
On the easiest pair (DF $\to$ AQ20), DINOv2 LP achieved 0.84 vs.\ EffNet LP at 0.60 and RN50 LP at 0.39.
All models showed a relative F1 drop of 44--55\% from their within-habitat (Protocol A) baselines, with CLIP LP showing the largest relative drop (55\%) and DINOv2 VPT the smallest (44\%).
In practical terms, foundation model features encode species-relevant structure that transfers across geographic regions, whereas CNN features are more tightly coupled to the training domain.

For monitoring programmes planning to deploy recognition at sites with no local training data, these results quantify what to expect: foundation models provide a useful starting point for within-taxon transfer (e.g., fish to fish), but cross-taxon transfer remains challenging and motivates the few-shot adaptation explored in Protocol~C.

\subsection{How Many Labels Do Ecologists Need? The Few-Shot Answer}

\begin{figure}[!htbp]
\centering
\includegraphics[width=0.92\textwidth]{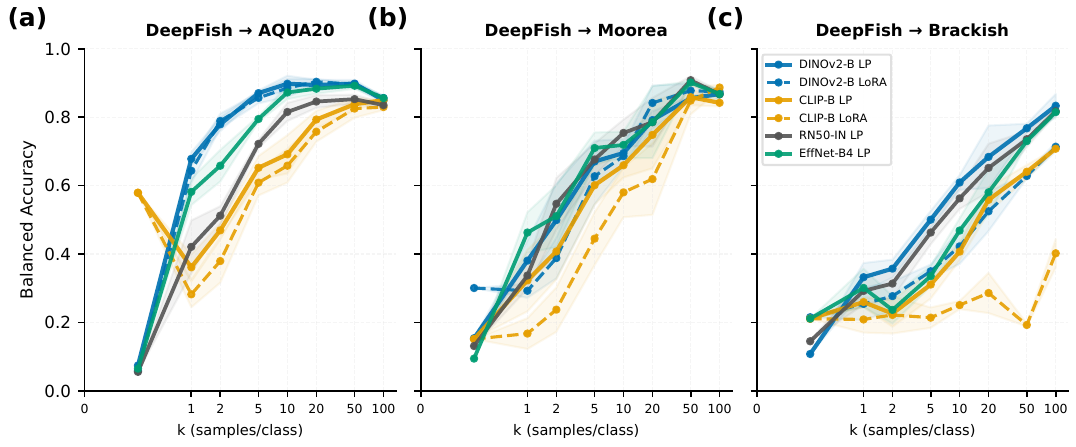}
\caption{Few-shot adaptation curves (Protocol C). Three panels correspond to transfer scenarios of increasing difficulty: DeepFish to AQUA20 (left), Moorea (centre), and Brackish (right). Balanced accuracy is plotted for six model--adaptation combinations; shaded regions indicate $\pm$1 standard deviation across five sampling trials ($k \leq 20$) or three trials ($k \geq 50$). DINOv2 LP reaches near-maximum performance by $k = 5$--$10$ on all transfers. CLIP zero-shot ($k = 0$) provides a useful starting point on AQUA20 but not on harder transfers.}
\label{fig:fewshot_curves}
\end{figure}

Protocol C directly addresses the title question.
Figure~\ref{fig:fewshot_curves} shows how recognition performance (balanced accuracy) scales with the number of labelled target samples per class ($k$) across three transfer scenarios of increasing difficulty (see Table~\ref{tab:fewshot}).
We report balanced accuracy rather than macro F1 because the few-shot sampling protocol progressively depletes the query set at high $k$, causing macro F1 to decline artifactually as minority classes are exhausted from evaluation.

Six model--adaptation combinations were evaluated: DINOv2 LP and LoRA, CLIP LP and LoRA, ResNet50 LP, and EfficientNet-B4 LP.
On the easiest transfer (DeepFish $\to$ AQUA20), DINOv2 LP exhibited a characteristic elbow: balanced accuracy rose steeply from $k = 0$ (0.07) to $k = 5$ (0.87), reaching $0.90$ at $k = 10$ and plateauing thereafter.
DINOv2 LoRA followed a similar trajectory (0.86 at $k = 5$, 0.90 at $k = 20$), confirming that the frozen features already capture transferable structure.

CLIP zero-shot ($k = 0$) provided a useful baseline that requires no labelled data.
Using text prompts alone, CLIP achieved balanced accuracy of 0.58 on AQUA20 (where class names align well with CLIP's language understanding), but only 0.15 on Moorea and 0.21 on Brackish, where specialised taxonomic labels fall outside CLIP's pretraining vocabulary.
While limited, this zero-shot capability could serve as a preliminary survey tool to prioritise which species or sites warrant further labelling effort.

For Moorea (the hardest Protocol C transfer, cross-taxon fish to coral), all models converged to balanced accuracy $\sim$0.84--0.89 at $k = 100$, though DINOv2 LP reached 80\% of its maximum by $k = 10$.
Brackish showed intermediate difficulty with wider separation between models: DINOv2 LP reached 0.83 at $k = 100$, while CLIP LP lagged at 0.71 and CLIP LoRA achieved only 0.40, suggesting that DINOv2 features transfer more effectively across biomes.

\begin{figure}[!htbp]
\centering
\includegraphics[width=0.92\textwidth]{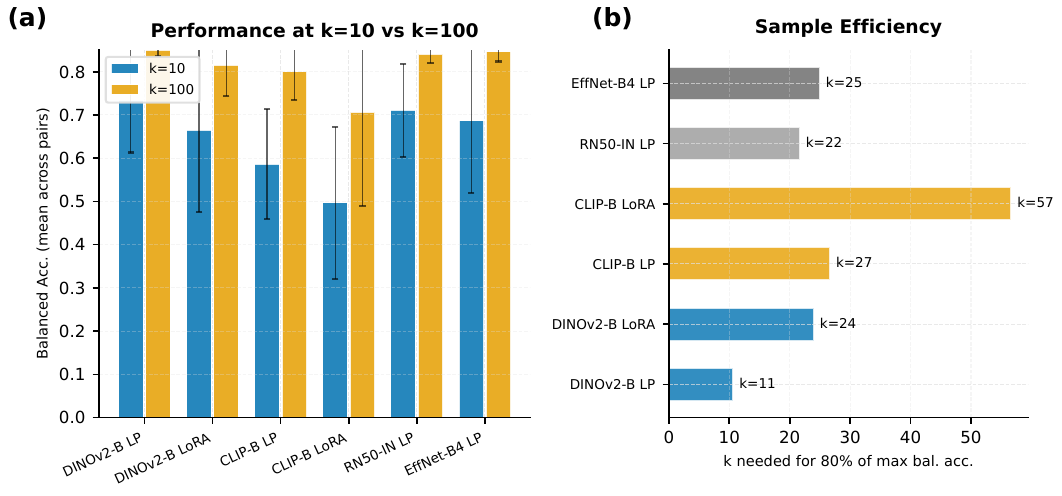}
\caption{Few-shot strategy analysis (Protocol C). (a) Balanced accuracy at $k = 10$ vs.\ $k = 100$ for each model--adaptation combination, averaged across transfer pairs; error bars show standard deviation across pairs. (b) Sample efficiency: the number of labelled samples per class ($k$) needed to reach 80\% of maximum balanced accuracy. DINOv2 LP reaches this threshold at $k \approx 11$ (interpolated between $k = 10$ and $k = 20$); 
}
\label{fig:fewshot_strategy}
\end{figure}

The practical message is that even a small number of labelled target samples (5--10 per class) can substantially boost performance over zero-shot transfer (Figure~\ref{fig:fewshot_strategy}).
DINOv2 LP reached 80\% of its maximum balanced accuracy at $k \approx 11$ (interpolated between tested values $k = 10$ and $k = 20$), the most sample-efficient combination tested.
This sample efficiency advantage of DINOv2 features, achieving strong performance with minimal labelling, directly lowers the barrier to deploying automated monitoring at new marine sites.

\begin{table}[t]
\centering
\caption{Few-shot headline numbers (Protocol C). Balanced accuracy $\pm$ standard deviation across sampling trials, averaged over three transfer pairs. Bold indicates the best result per column.}
\label{tab:fewshot}
\small
\begin{tabular}{@{}lcccccccc@{}}
\toprule
Model--Adaptation & $k=0$ & $k=1$ & $k=2$ & $k=5$ & $k=10$ & $k=20$ & $k=50$ & $k=100$ \\
\midrule
DINOv2-B LP       & .11 & \textbf{.46}$\pm$.15 & \textbf{.55}$\pm$.18 & \textbf{.68}$\pm$.15 & \textbf{.73}$\pm$.12 & \textbf{.79}$\pm$.09 & \textbf{.84}$\pm$.05 & \textbf{.85}$\pm$.01 \\
DINOv2-B LoRA     & .19 & .40$\pm$.18          & .48$\pm$.22          & .61$\pm$.21          & .66$\pm$.19          & .76$\pm$.17          & .80$\pm$.12          & .81$\pm$.07 \\
CLIP-B LP         & \textbf{.31} & .31$\pm$.04   & .37$\pm$.10          & .52$\pm$.15          & .59$\pm$.13          & .70$\pm$.10          & .78$\pm$.10          & .80$\pm$.07 \\
CLIP-B LoRA       & \textbf{.31} & .22$\pm$.05   & .28$\pm$.07          & .42$\pm$.16          & .50$\pm$.18          & .55$\pm$.20          & .62$\pm$.30          & .71$\pm$.22 \\
RN50-IN LP        & .11 & .35$\pm$.05          & .46$\pm$.10          & .62$\pm$.11          & .71$\pm$.11          & .76$\pm$.08          & .83$\pm$.07          & .84$\pm$.02 \\
EffNet-B4 LP      & .12 & .45$\pm$.12          & .47$\pm$.17          & .61$\pm$.20          & .69$\pm$.17          & .75$\pm$.13          & .84$\pm$.08          & .85$\pm$.02 \\
\bottomrule
\end{tabular}
\end{table}

\subsection{What Makes Foundation Models Different? Feature Analysis}

To understand why foundation models generalise across habitats while CNN baselines do not, we analysed the learned feature representations using t-SNE \citep{vandermaaten2008tsne}, a technique that projects a model's high-dimensional internal representations onto a two-dimensional plot so that images the model treats as similar appear close together.

\subsubsection{Feature Space Structure}

\begin{figure}[!htbp]
\centering
\includegraphics[width=0.92\textwidth]{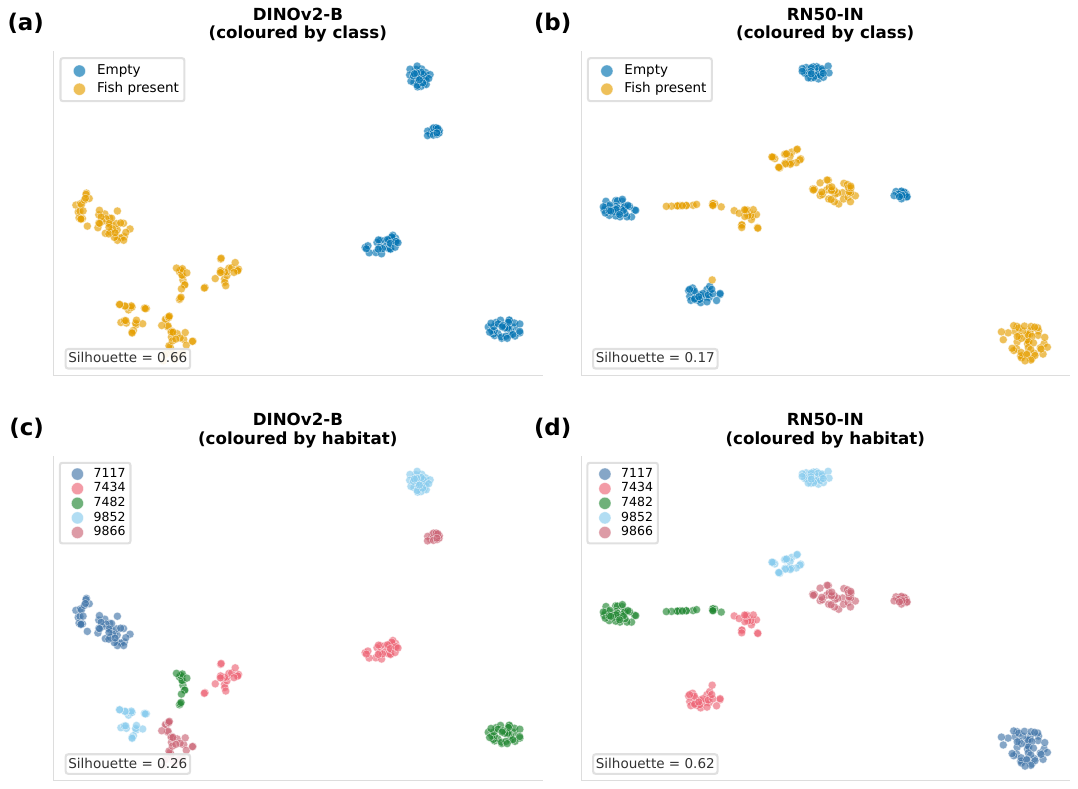}
\caption{t-SNE visualisation of learned features for 250 DeepFish images across five habitats. Top row: features coloured by class (empty vs.\ fish present); bottom row: same features coloured by habitat of origin. Silhouette scores (Sil) quantify clustering strength. DINOv2 + LoRA (left) achieves strong class separation (Sil\,=\,0.66) with habitats fully mixed (Sil\,=\,0.26), indicating habitat-invariant representations. ResNet-50 + full fine-tuning (right) shows weak class separation (Sil\,=\,0.17) but strong habitat clustering (Sil\,=\,0.62), indicating reliance on habitat-specific visual shortcuts.}
\label{fig:tsne}
\end{figure}

Figure~\ref{fig:tsne} presents t-SNE projections of features extracted from 250 DeepFish test images spanning five habitats, coloured by class label (top row) and by habitat of origin (bottom row).
DINOv2 + LoRA features cluster tightly by class (empty vs.\ fish present), achieving a silhouette score of 0.66, a measure of cluster quality ranging from $-1$ to $1$, where higher values indicate tighter, better-separated groups, with samples from different habitats intermixed within each class cluster (habitat silhouette = 0.26).
This indicates that the model has learned class-diagnostic representations that are invariant to habitat-specific visual conditions.
ResNet-50 features, by contrast, show weak class separation (silhouette = 0.17) but strong habitat clustering (silhouette = 0.62): samples from the same site group together regardless of class identity, suggesting that the model relies on background cues (substrate colour, water clarity) rather than species-diagnostic morphology.

This difference has direct ecological implications.
A model that clusters by species will generalise to new habitats because it has learned what a fish looks like regardless of background.
A model that clusters by habitat will fail at new sites because it has learned what a particular site looks like, not what the organisms within it look like.

\subsubsection{Efficiency Frontier}

Appendix~S1: Figure~S2 plots the performance--efficiency trade-off for all model--adaptation combinations, with trainable parameters on the horizontal axis (log scale) and macro F1 on the vertical axis.
DINOv2-Base linear probe is the sole Pareto-optimal point, achieving the highest F1 (0.70) with only 1{,}538 trainable parameters, four orders of magnitude fewer than LoRA (149K) or full fine-tuning (17--24M).
This efficiency has practical relevance: linear probing of DINOv2-Base requires minimal GPU memory and trains in minutes, making foundation model performance accessible to ecology research groups with limited computational resources.

\section{Discussion}

\subsection{Practical Recommendations for Ecologists}

The results across all three protocols converge on a set of concrete recommendations for deploying automated species recognition at new marine sites.
We present these as a decision framework based on the number of labelled images available from the target site:

\begin{itemize}
    \item \textbf{Zero labels:} Use CLIP zero-shot classification with species-descriptive text prompts. This requires no labelled data and provides a preliminary estimate of species composition, useful for survey prioritisation. Expected performance: balanced accuracy = 0.15--0.58, depending on how well species names align with CLIP's vocabulary and target domain similarity.
    \item \textbf{1--5 labels per species:} Use DINOv2-Base with a linear probe on the frozen features. With very few labels, the pretrained features are already informative and the risk of overfitting is minimised. Expected performance: balanced accuracy rises steeply from 0.46 at $k=1$ to 0.68 at $k=5$ (averaged across transfer pairs).
    \item \textbf{10--20 labels per species (recommended):} Use DINOv2-Base with a linear probe. This is the practical sweet spot: 10 labelled images per species (an annotation effort of 1--4 hours depending on species count) achieves near-plateau performance. Expected performance: balanced accuracy $\approx$ 0.73 at $k=10$, rising to 0.79 at $k=20$ (averaged across transfer pairs).
    \item \textbf{50+ labels per species:} Marginal gains above $k=50$ are small (balanced accuracy 0.84 at $k=50$, 0.85 at $k=100$). LoRA adaptation and full fine-tuning of CNNs offer no clear advantage over a frozen linear probe at any $k$ value, while requiring substantially more computation.
\end{itemize}

\textbf{Hardware requirements.} Linear probing of DINOv2-Base requires minimal GPU memory and trains in minutes on a single consumer GPU. Even LoRA adaptation runs on a GPU with 16\,GB of memory and completes in under one hour. Full fine-tuning of the same backbone requires 40\,GB or more. Since the best-performing configuration (frozen linear probe) is also the most efficient, foundation model performance is accessible to research groups without specialised computing infrastructure.

\subsection{Why Foundation Models Succeed Underwater}

The feature analysis (Section~3.4) reveals the mechanism behind foundation model transferability: self-supervised pretraining produces representations that encode structural and textural properties of objects, suppressing domain-specific confounds such as water colour and substrate type.

The self-distillation objective used in DINOv2 training is instructive.
By learning to produce consistent representations across different augmented views of the same image (which include colour jitter, random cropping, and geometric distortion), the model is explicitly trained to ignore surface-level visual variation and encode deeper structural invariances.
This is precisely what is needed for cross-habitat transfer in marine systems, where the same species appears under dramatically different colour profiles, lighting conditions, and backgrounds.

The attention mechanism in Vision Transformers plays a complementary role.
Unlike CNNs, which process images through fixed local receptive fields, transformers can attend selectively to any spatial location.
Prior work has shown that self-supervised ViTs learn to attend to salient foreground objects \citep{caron2021dino}, suggesting that this flexibility enables the model to focus on organisms and suppress background variation, much as an expert taxonomist examines diagnostic features (body shape, fin morphology, skeletal structure) rather than the surrounding environment.
CNNs, constrained to local feature extraction, are more susceptible to encoding spatially co-occurring habitat features alongside target features.

\subsection{Limitations, Uncertainty, and Transferability}

Several limitations bound the scope of our conclusions.

\textbf{Benchmark composition.} All five datasets are publicly available benchmarks collected under specific conditions. Real-world deployment may encounter challenges not represented here, including novel species absent from training data, extreme turbidity or biofouling on camera housings, and camera artefacts specific to particular platforms. Our difficulty gradient is observational rather than experimentally controlled. Geographic distance, taxonomic distance, number of target classes (2 to 14), and class imbalance severity are correlated across our benchmark and cannot be cleanly disentangled. The near-chance performance on the Coralscapes transfer (F1 $\approx$ 0.04) reflects not only the ecological distance between fish and coral but also the challenge of nearest-prototype matching into 14 severely imbalanced classes. We note, however, that these confounds reflect ecological reality: harder real-world transfers typically involve more classes, greater imbalance, and larger visual domain gaps simultaneously. The gradient should therefore be interpreted as an ordinal ranking of practical transfer difficulty rather than a controlled measurement of any single factor.

\textbf{Task simplification.} DeepFish and Brackish use binary or coarse species labels. Fine-grained species identification (distinguishing closely related congeners) is a harder task that may not benefit as strongly from pretrained features. The conversion from detection and segmentation annotations to image-level classification labels, while enabling cross-task comparison, discards spatial information that may be ecologically relevant.

\textbf{Temporal considerations.} DeepFish and Brackish contain frames extracted from video. Although we split by video to mitigate temporal data leakage, consecutive frames within a video share visual context, and our within-video independence assumption may be imperfect. Video-specific temporal models are not evaluated.

\textbf{Transfer directionality.} All cross-dataset transfers in Protocols B and C use DeepFish as the sole source dataset. We did not evaluate other transfer directions (e.g., coral-to-coral or coral-to-fish), and transferability from non-fish sources remains untested.

\textbf{Ecosystem coverage.} Our benchmark spans tropical to temperate marine systems but does not include freshwater, deep-sea, pelagic, or polar environments. Transferability to these domains remains an open question, though the underlying methodology (foundation model plus parameter-efficient adaptation) is domain-agnostic.

\textbf{Computational costs.} While LoRA adaptation is efficient (16\,GB GPU memory, under one hour training), the DINOv2-Base backbone itself contains 86 million parameters and requires GPU inference. For CLIP configurations, the text encoder (63.4M parameters) is also loaded in memory even though it is not trained, increasing the total memory footprint to approximately 149M parameters. For resource-constrained settings (e.g., on-board processing on autonomous underwater vehicles), model distillation or quantisation may be necessary, which we do not evaluate.

\subsection{Future Directions}

Several extensions would strengthen the practical value of this framework.
Active learning methods that use model uncertainty to prioritise which images an expert should label could further reduce annotation requirements by selecting the most informative samples \citep{tuia2022perspectives}.
Video-aware models that leverage temporal continuity (tracking individuals across frames) could improve both accuracy and ecological utility by linking detections to abundance estimates.
Foundation models pretrained specifically on underwater imagery, as underwater image repositories grow, may provide even stronger starting representations for marine tasks.
Multi-task learning that jointly addresses classification, detection, and segmentation could unify the currently fragmented landscape of marine computer vision tools.
Finally, integration with ecological modelling (species distribution models, community composition analysis) would close the loop from raw imagery to ecological inference.


\section{Conclusions}

We present a decision framework, validated across five marine ecosystems spanning three oceans and three taxonomic groups, that allows ecological monitoring programmes to budget labelling effort against expected recognition accuracy when deploying automated species recognition at new sites.
The framework rests on three findings, established through 928 systematic experimental runs across within-habitat, cross-dataset, and few-shot adaptation protocols.
First, as few as 10--20 labelled images per species suffice to deploy reliable recognition at a new site (an annotation effort of 1--4 hours), reducing the labelling bottleneck that currently constrains the geographic scope of automated monitoring by roughly an order of magnitude.
Second, frozen self-supervised foundation model features (DINOv2 with a linear classifier) match or exceed the performance of fully fine-tuned convolutional baselines across every transfer scenario tested, while requiring four orders of magnitude fewer trainable parameters and no specialised computing infrastructure.
Third, foundation models generalise where convolutional baselines fail because they encode species-diagnostic representations that are habitat-invariant rather than habitat-specific visual shortcuts.

For monitoring programmes, the practical implication is that expanding to a new reef, estuary, or kelp forest no longer requires restarting the labelling cycle from scratch.
A modest annotation effort, guided by the framework presented here, is sufficient to achieve reliable performance.
Although demonstrated on marine imagery, the evaluation methodology and decision rules apply to any ecological monitoring system that must generalise across sites with limited labelled data, from camera trap networks to acoustic surveys to aerial monitoring.


\section*{Data Availability Statement}

All code, configurations, and analysis scripts for reproducing the experiments are publicly available at \url{https://github.com/alzayats/cross-habitat-marine}.
Upon acceptance, a versioned snapshot of the code together with all derived analysis outputs (per-run metrics, feature embeddings, and figure-generation scripts) will be archived in the Dryad Digital Repository with a citable DOI.
The five datasets used in this study are publicly available and were obtained from their original sources: DeepFish \citep{saleh2020deepfish}, AQUA20 \citep{fuad2025aqua20}, Moorea Labeled Corals \citep{beijbom2012moorea, edmunds2019mcrlter}, Coralscapes \citep{sauder2025coralscapes}, and Brackish \citep{pedersen2019brackish}.


\section*{CO$_2$ Emission Related to Experiments}

All experiments were conducted on a single NVIDIA GeForce RTX 4090 GPU (450\,W TDP) at James Cook University's High Performance Computing facilities.
The experiments comprised 240 training runs (Protocol~A), 40 evaluation runs (Protocol~B), and 648 training and evaluation runs (Protocol~C), for a total of approximately 730 GPU hours.
Assuming 80\% average GPU utilisation, a power usage effectiveness (PUE) of 1.1, and the Australian average grid carbon intensity of 0.65\,kgCO$_2$/kWh (2023 estimate), we estimate total energy consumption at approximately 289\,kWh and associated CO$_2$ emissions at approximately 188\,kg.
For context, this is comparable to approximately 1{,}200\,km of driving in an average passenger car.
We note that the most parameter-efficient configuration (DINOv2-Base linear probe, 1{,}538 trainable parameters) trains in under 5 minutes per habitat, suggesting that the majority of compute was spent on the full factorial comparison rather than on the recommended deployment configuration.


\section*{Acknowledgements}

We thank the providers of the five public datasets used in this study (DeepFish, AQUA20, Moorea Labeled Corals, Coralscapes, and Brackish), whose openly released imagery made this benchmark possible; full citations and access details are given in the Data Availability Statement. Computational resources were provided by James Cook University's High Performance Computing facilities (Townsville, QLD, Australia).

\textbf{Use of generative AI.} Generative AI tools (Anthropic Claude) were used to assist with manuscript drafting (text refinement and clarity edits) and with code documentation. The AI tools did not contribute to research design, dataset curation, model implementation, statistical analyses, or scientific interpretation of results. All AI-assisted text was reviewed and revised by the authors, and all AI-assisted code was inspected, tested, and integrated into the publicly archived repository by the authors. The corresponding author takes full responsibility for the accuracy of all content.

\section*{Funding}

This research did not receive any specific grant from funding agencies in the public, commercial, or not-for-profit sectors.


\section*{Author Contributions}

Alzayat Saleh: Conceptualization, methodology, software, data curation, formal analysis, investigation, writing (original draft), visualization, and project administration.
Mostafa Rahimi Azghadi: Conceptualization and review \& editing of the manuscript.

\section*{Conflict of Interest Statement}

The authors declare no conflicts of interest. No financial, professional, or personal relationships have influenced the design, conduct, analysis, or reporting of this work.


\bibliography{references}

@inproceedings{caron2021dino,
  title={Emerging Properties in Self-Supervised Vision Transformers},
  author={Caron, Mathilde and Touvron, Hugo and Misra, Ishan and J{\'e}gou, Herv{\'e} and Mairal, Julien and Bojanowski, Piotr and Joulin, Armand},
  booktitle={Proceedings of the IEEE/CVF International Conference on Computer Vision},
  pages={9630--9640},
  year={2021}
}

@article{oquab2024dinov2,
  title={{DINOv2}: Learning Robust Visual Features without Supervision},
  author={Oquab, Maxime and Darcet, Timoth{\'e}e and Moutakanni, Th{\'e}o and Vo, Huy and Szafraniec, Marc and Khalidov, Vasil and Fernandez, Pierre and Haziza, Daniel and Massa, Francisco and El-Nouby, Alaaeldin and others},
  journal={Transactions on Machine Learning Research},
  year={2024}
}

@inproceedings{radford2021clip,
  title={Learning Transferable Visual Models from Natural Language Supervision},
  author={Radford, Alec and Kim, Jong Wook and Hallacy, Chris and Ramesh, Aditya and Goh, Gabriel and Agarwal, Sandhini and Sastry, Girish and Askell, Amanda and Mishkin, Pamela and Clark, Jack and others},
  booktitle={International Conference on Machine Learning},
  pages={8748--8763},
  year={2021},
  organization={PMLR}
}

@inproceedings{he2016resnet,
  title={Deep Residual Learning for Image Recognition},
  author={He, Kaiming and Zhang, Xiangyu and Ren, Shaoqing and Sun, Jian},
  booktitle={Proceedings of the IEEE Conference on Computer Vision and Pattern Recognition},
  pages={770--778},
  year={2016}
}

@inproceedings{tan2019efficientnet,
  title={{EfficientNet}: Rethinking Model Scaling for Convolutional Neural Networks},
  author={Tan, Mingxing and Le, Quoc},
  booktitle={International Conference on Machine Learning},
  pages={6105--6114},
  year={2019},
  organization={PMLR}
}

@inproceedings{hu2022lora,
  title={{LoRA}: Low-Rank Adaptation of Large Language Models},
  author={Hu, Edward J and Shen, Yelong and Wallis, Phillip and Allen-Zhu, Zeyuan and Li, Yuanzhi and Wang, Shean and Wang, Lu and Chen, Weizhu},
  booktitle={International Conference on Learning Representations},
  year={2022}
}

@inproceedings{jia2022vpt,
  title={Visual Prompt Tuning},
  author={Jia, Menglin and Tang, Luming and Chen, Bor-Chun and Cardie, Claire and Belongie, Serge and Hariharan, Bharath and Lim, Ser-Nam},
  booktitle={European Conference on Computer Vision},
  pages={709--727},
  year={2022},
  organization={Springer}
}

@inproceedings{loshchilov2019adamw,
  title={Decoupled Weight Decay Regularization},
  author={Loshchilov, Ilya and Hutter, Frank},
  booktitle={International Conference on Learning Representations},
  year={2019}
}

@article{fuad2025aqua20,
  title={{AQUA20}: A Benchmark Dataset for Underwater Species Classification under Challenging Conditions},
  author={Fuad, Taufikur Rahman and Ahmed, Sabbir and Ivan, Shahriar},
  journal={arXiv preprint arXiv:2506.17455},
  year={2025}
}

@article{saleh2020deepfish,
  title={A Realistic Fish-Habitat Dataset to Evaluate Algorithms for Underwater Visual Analysis},
  author={Saleh, Alzayat and Laradji, Issam H and Konovalov, Dmitry A and Bradley, Michael and Vazquez, David and Sheaves, Marcus},
  journal={Scientific Reports},
  volume={10},
  number={1},
  pages={14671},
  year={2020},
  publisher={Nature Publishing Group}
}

@inproceedings{pedersen2019brackish,
  title={Detection of Marine Animals in a New Underwater Dataset with Varying Visibility},
  author={Pedersen, Malte and Haurum, Joakim Bruslund and Gade, Rikke and Moeslund, Thomas B and Madsen, Niels},
  booktitle={Proceedings of the IEEE/CVF Conference on Computer Vision and Pattern Recognition Workshops},
  pages={18--26},
  year={2019}
}

@inproceedings{beijbom2012moorea,
  title={Automated Annotation of Coral Reef Survey Images},
  author={Beijbom, Oscar and Edmunds, Peter J and Kline, David I and Mitchell, B Greg and Kriegman, David},
  booktitle={Proceedings of the IEEE Conference on Computer Vision and Pattern Recognition},
  pages={1170--1177},
  year={2012}
}

@inproceedings{sauder2025coralscapes,
  title={The Coralscapes Dataset: Semantic Scene Understanding in Coral Reefs},
  author={Sauder, Jonathan and Domazetoski, Viktor and Banc-Prandi, Guilhem and Perna, Gabriela and Meibom, Anders and Tuia, Devis},
  booktitle={Proceedings of the IEEE/CVF International Conference on Computer Vision Workshops},
  year={2025}
}

@misc{edmunds2019mcrlter,
  title={{MCR LTER}: Coral Reef: Computer Vision: {M}oorea Labeled Corals},
  author={{Moorea Coral Reef LTER} and Edmunds, Peter J},
  year={2019},
  publisher={Environmental Data Initiative},
  doi={10.6073/pasta/88dde0e68ab5232a470389f4bedd1892},
  note={ver 3, knb-lter-mcr.5006.3}
}

@article{weinstein2018computer,
  title={A Computer Vision for Animal Ecology},
  author={Weinstein, Ben G},
  journal={Journal of Animal Ecology},
  volume={87},
  number={3},
  pages={533--545},
  year={2018},
  publisher={Wiley}
}

@article{christin2019applications,
  title={Applications for Deep Learning in Ecology},
  author={Christin, Sylvain and Hervet, {\'E}ric and Lecomte, Nicolas},
  journal={Methods in Ecology and Evolution},
  volume={10},
  number={10},
  pages={1632--1644},
  year={2019},
  publisher={Wiley}
}

@article{ditria2020automating,
  title={Automating the Analysis of Fish Abundance Using Object Detection: Optimizing Animal Ecology with Deep Learning},
  author={Ditria, Ellen M and Lopez-Marcano, Sebastian and Sievers, Michael and Jinks, Eric L and Brown, Christopher J and Connolly, Rod M},
  journal={Frontiers in Marine Science},
  volume={7},
  pages={429},
  year={2020},
  publisher={Frontiers}
}

@article{gonzalez2020monitoring,
  title={Monitoring of Coral Reefs Using Artificial Intelligence: A Feasible and Cost-Effective Approach},
  author={Gonz{\'a}lez-Rivero, Manuel and Beijbom, Oscar and Rodriguez-Ramirez, Alberto and Bryant, Dominic E P and Ganase, Anjani and Gonz{\'a}lez-Marrero, Yeray and Herrera-Reveles, Ana and Kennedy, Emma V and Kim, Catherine J S and Lopez-Marcano, Sebastian and Markey, Kathryn and Neal, Benjamin P and Osborne, Kate and Reyes-Nivia, Catalina and Sampayo, Eugenia M and Stolberg, Kristin and Taylor, Abbie and Vercelloni, Julie and Wyatt, Mathew and Hoegh-Guldberg, Ove},
  journal={Remote Sensing},
  volume={12},
  number={3},
  pages={489},
  year={2020},
  publisher={MDPI}
}

@article{villon2018coral,
  title={A Deep Learning Method for Accurate and Fast Identification of Coral Reef Fishes in Underwater Images},
  author={Villon, S{\'e}bastien and Mouillot, David and Chaumont, Marc and Darling, Emily S and Subsol, G{\'e}rard and Claverie, Thomas and Vill{\'e}ger, S{\'e}bastien},
  journal={Ecological Informatics},
  volume={48},
  pages={238--244},
  year={2018},
  publisher={Elsevier}
}

@article{tuia2022perspectives,
  title={Perspectives in Machine Learning for Wildlife Conservation},
  author={Tuia, Devis and Kellenberger, Benjamin and Beery, Sara and Costelloe, Blair R and Zuffi, Silvia and Risse, Benjamin and Mathis, Alexander and Mathis, Mackenzie W and van Langevelde, Frank and Burghardt, Tilo and others},
  journal={Nature Communications},
  volume={13},
  number={1},
  pages={792},
  year={2022},
  publisher={Nature Publishing Group}
}

@article{schneider2019past,
  title={Past, Present and Future Approaches Using Computer Vision for Animal Re-Identification from Camera Trap Data},
  author={Schneider, Stefan and Taylor, Graham W and Linquist, Stefan and Kremer, Stefan C},
  journal={Methods in Ecology and Evolution},
  volume={10},
  number={4},
  pages={461--470},
  year={2019},
  publisher={Wiley}
}

@inproceedings{stevens2024bioclip,
  title={{BioCLIP}: A Vision Foundation Model for the Tree of Life},
  author={Stevens, Samuel and Wu, Jiaman and Thompson, Matthew J and Campolongo, Elizabeth G and Song, Chan Hee and Carlyn, David Edward and Dong, Li and Dahdul, Wasila M and Stewart, Charles and Berger-Wolf, Tanya and Chao, Wei-Lun and Su, Yu},
  booktitle={Proceedings of the IEEE/CVF Conference on Computer Vision and Pattern Recognition},
  year={2024}
}

@article{katija2022fathomnet,
  title={{FathomNet}: A Global Image Database for Enabling Artificial Intelligence in the Ocean},
  author={Katija, Kakani and Orenstein, Eric and Schlining, Brian and Lundsten, Lonny and Barnard, Kevin and Sainz, Giovanna and Boulais, Oceane and Cromwell, Megan and Butler, Erin and Woodward, Benjamin and Bell, Katherine L},
  journal={Scientific Reports},
  volume={12},
  number={1},
  pages={15914},
  year={2022},
  publisher={Nature Publishing Group}
}

@article{saleh2022fish_survey,
  title={Computer Vision and Deep Learning for Fish Classification in Underwater Habitats: A Survey},
  author={Saleh, Alzayat and Sheaves, Marcus and Rahimi Azghadi, Mostafa},
  journal={Fish and Fisheries},
  volume={23},
  number={4},
  pages={977--999},
  year={2022},
  publisher={Wiley}
}

@article{vandermaaten2008tsne,
  title={Visualizing Data using {t-SNE}},
  author={Van der Maaten, Laurens and Hinton, Geoffrey},
  journal={Journal of Machine Learning Research},
  volume={9},
  number={86},
  pages={2579--2605},
  year={2008}
}

@article{durden2016perspectives,
  title={Perspectives in Visual Imaging for Marine Biology and Ecology: From Acquisition to Understanding},
  author={Durden, Jennifer M and Schoening, Timm and Althaus, Franziska and Friedman, Ariell and Garcia, Rafael and Glover, Adrian G and Greinert, Jens and Jacobsen Stout, Nancy and Jones, Daniel OB and Jordt, Anne and others},
  journal={Oceanography and Marine Biology: An Annual Review},
  volume={54},
  pages={1--72},
  year={2016}
}

@article{mallet2014underwater,
  title={Underwater Video Techniques for Observing Coastal Marine Biodiversity: A Review of Sixty Years of Publications (1952--2012)},
  author={Mallet, Delphine and Pelletier, Dominique},
  journal={Fisheries Research},
  volume={154},
  pages={44--62},
  year={2014},
  publisher={Elsevier}
}

@article{thompson1997interobserver,
  title={Observer Effects and Training in Underwater Visual Surveys of Reef Fishes},
  author={Thompson, Andrew A and Mapstone, Bruce D},
  journal={Marine Ecology Progress Series},
  volume={154},
  pages={53--63},
  year={1997},
  publisher={Inter-Research}
}

@article{beijbom2015coralnet,
  title={Towards Automated Annotation of Benthic Survey Images: Variability of Human Experts and Operational Modes of Automation},
  author={Beijbom, Oscar and Edmunds, Peter J and Roelfsema, Chris and Smith, Jennifer and Kline, David I and Neal, Benjamin P and Dunlap, Matthew J and Moriarty, Vincent and Fan, Tung-Yung and Tan, Chih-Jui and others},
  journal={PLoS ONE},
  volume={10},
  number={7},
  pages={e0130312},
  year={2015},
  publisher={Public Library of Science}
}

@article{williams2019leveraging,
  title={Leveraging Automated Image Analysis Tools to Transform Our Capacity to Assess Status and Trends of Coral Reefs},
  author={Williams, Ian D and Couch, Courtney S and Beijbom, Oscar and Oliver, Thomas A and Vargas-Angel, Bernardo and Schumacher, Brett D and Brainard, Russell E},
  journal={Frontiers in Marine Science},
  volume={6},
  pages={222},
  year={2019},
  publisher={Frontiers}
}

@inproceedings{deng2009imagenet,
  title={{ImageNet}: A Large-Scale Hierarchical Image Database},
  author={Deng, Jia and Dong, Wei and Socher, Richard and Li, Li-Jia and Li, Kai and Fei-Fei, Li},
  booktitle={Proceedings of the IEEE Conference on Computer Vision and Pattern Recognition},
  pages={248--255},
  year={2009}
}

\newpage
\appendix
\section{Supplementary Material} 




\thispagestyle{empty}

\begin{flushleft}
\textbf{Appendix S1.} \\[2pt]
Companion to: \emph{How many labels do you need? A decision framework for
cross-habitat marine species recognition.} \\[2pt]
Alzayat Saleh, Mostafa Rahimi Azghadi. \\[2pt]
Submitted to \emph{Ecological Applications}.
\end{flushleft}

\vspace{1.5em}

\noindent
This appendix provides supporting tables and figures referenced from the main
manuscript. Table~S1 reports exact trainable parameter counts for all ten
model--adaptation configurations evaluated in Protocol~A. Table~S2 reports
Protocol~A results with both macro F1 and balanced accuracy for cross-protocol
comparability. Figure~S1 shows training dynamics for the top six
model--adaptation combinations. Figure~S2 shows the efficiency--performance
trade-off across all combinations.

\vspace{1em}


\begin{table}[h]
\centering
\caption{Exact trainable parameter counts for all ten model--adaptation
configurations evaluated in Protocol~A. For CLIP configurations, the text
encoder (63.4M parameters) is loaded to support zero-shot classification but is
excluded from the training optimiser; trainable counts reflect only the visual
encoder and classification head.}
\label{tab:supp_params}
\small
\begin{tabular}{@{}llrrr@{}}
\toprule
Model & Adaptation & Total Params & Trainable Params & Trainable \% \\
\midrule
DINOv2-Base      & Linear Probe    & 86{,}582{,}018 & 1{,}538       & 0.002 \\
DINOv2-Base      & LoRA ($r=4$)    & 86{,}729{,}474 & 148{,}994     & 0.172 \\
DINOv2-Base      & VPT-Deep        & 86{,}674{,}178 & 93{,}698      & 0.108 \\
CLIP-Base        & Linear Probe    & 86{,}193{,}667 & 1{,}027       & 0.001 \\
CLIP-Base        & LoRA ($r=4$)    & 86{,}267{,}395 & 74{,}755      & 0.087 \\
CLIP-Base        & VPT-Deep        & 86{,}285{,}827 & 93{,}187      & 0.108 \\
ResNet-50        & Linear Probe    & 23{,}512{,}130 & 4{,}098       & 0.017 \\
ResNet-50        & Full Fine-tune  & 23{,}512{,}130 & 23{,}512{,}130 & 100.0 \\
EfficientNet-B4  & Linear Probe    & 17{,}552{,}202 & 3{,}586       & 0.020 \\
EfficientNet-B4  & Full Fine-tune  & 17{,}552{,}202 & 17{,}552{,}202 & 100.0 \\
\bottomrule
\end{tabular}
\end{table}


\begin{table}[h]
\centering
\caption{Protocol~A results reported with both macro F1 and balanced accuracy
for cross-protocol comparability. Values are mean $\pm$ SD across 20 habitats
(seed 42). Rankings are consistent across both metrics.}
\label{tab:supp_metrics}
\small
\begin{tabular}{@{}lcc@{}}
\toprule
Configuration & Macro F1 & Balanced Accuracy \\
\midrule
DINOv2-B LP       & $0.70 \pm 0.28$ & $0.75 \pm 0.24$ \\
DINOv2-B LoRA     & $0.64 \pm 0.30$ & $0.72 \pm 0.23$ \\
DINOv2-B VPT      & $0.65 \pm 0.30$ & $0.71 \pm 0.24$ \\
CLIP-B LP         & $0.60 \pm 0.23$ & $0.67 \pm 0.16$ \\
CLIP-B LoRA       & $0.58 \pm 0.23$ & $0.65 \pm 0.18$ \\
CLIP-B VPT        & $0.59 \pm 0.23$ & $0.66 \pm 0.18$ \\
EffNet-B4 FT      & $0.67 \pm 0.24$ & $0.72 \pm 0.21$ \\
EffNet-B4 LP      & $0.57 \pm 0.22$ & $0.63 \pm 0.18$ \\
RN50 FT           & $0.63 \pm 0.27$ & $0.69 \pm 0.22$ \\
RN50 LP           & $0.56 \pm 0.23$ & $0.62 \pm 0.19$ \\
\bottomrule
\end{tabular}
\end{table}


\begin{figure}[h]
\centering
\includegraphics[width=0.92\textwidth]{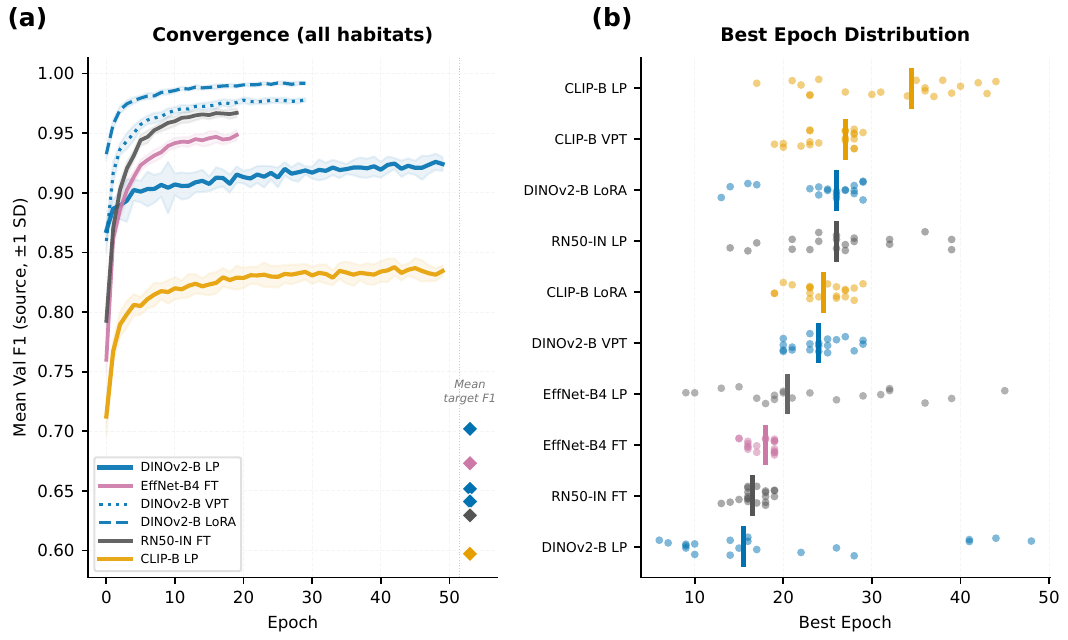}
\caption{Training dynamics (Protocol A). (a) Convergence curves showing mean
validation F1 ($\pm$ 1 SD) over epochs, averaged across all habitats, for the
top six model--adaptation combinations. Diamond markers on the right show mean
target test F1. DINOv2 linear probe converges in 2--3 epochs. (b) Distribution
of best epoch across all runs per combination; median lines overlaid.}
\label{fig:training_dynamics}
\end{figure}


\begin{figure}[h]
\centering
\includegraphics[width=0.92\textwidth]{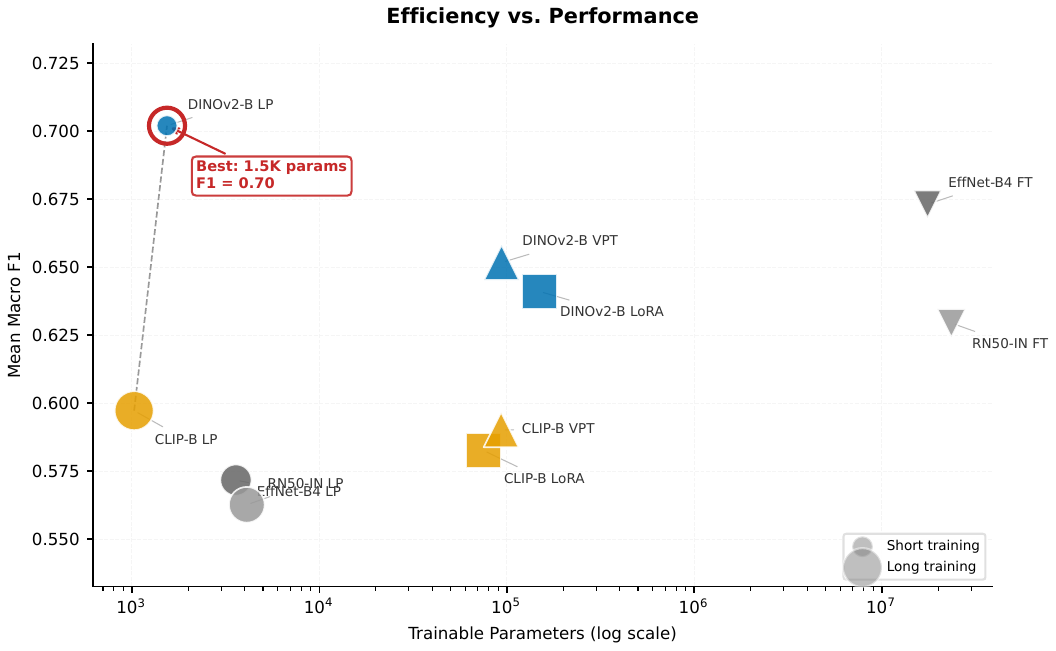}
\caption{Efficiency--performance trade-off (Protocol A). Each point represents
one model--adaptation combination; the $x$-axis shows trainable parameters
(log scale) and the $y$-axis shows mean macro F1 across all habitats. Bubble
size encodes training time. DINOv2-Base linear probe (circled) achieves the
highest F1 with only 1{,}538 trainable parameters, dominating the Pareto
frontier at four orders of magnitude fewer parameters than full fine-tuning
alternatives.}
\label{fig:pareto_within}
\end{figure}


\end{document}